\documentclass[review,10.8pt]{elsarticle}

\usepackage{hyperref}
\usepackage[utf8]{inputenc} %unicode support
\usepackage{graphicx}
\usepackage{subfigure}
\usepackage{amsmath}
\usepackage{adjustbox}						
\usepackage{color}
\usepackage[normalem]{ulem}
\usepackage{multirow}
\usepackage{booktabs}
\usepackage{lineno}

\usepackage{soul}
\usepackage{todonotes}

\usepackage{lipsum}
\usepackage{natbib}

\journal{Expert Systems with Applications}

\biboptions{authoryear}
\begin{document}
\begin{frontmatter}

\title{Continuous Emotion Recognition with Spatiotemporal Convolutional Neural Networks}

\author{Thomas Teixeira}%\ead{thomas.teixeira.1@ens.etsmtl.ca}
\address{\'{E}cole de Technologie Sup\'{e}rieure, Universit\'{e} du Qu\'{e}bec\\ 1100, rue Notre-Dame Ouest, H3C 1K3, Montreal, QC, Canada\\
\url{thomas.teixeira.1@ens.etsmtl.ca}}

\author{Éric Granger}%\ead{eric.granger@etsmtl.ca}
\address{\'{E}cole de Technologie Sup\'{e}rieure, Universit\'{e} du Qu\'{e}bec\\ 1100, rue Notre-Dame Ouest, H3C 1K3, Montreal, QC, Canada\\
\url{eric.granger@etsmtl.ca}}

\author{Alessandro Lameiras Koerich\fnref{myfootnote}}%\ead{alessandro.koerich@etsmtl.ca}
\address{\'{E}cole de Technologie Sup\'{e}rieure, Universit\'{e} du Qu\'{e}bec\\ 1100, rue Notre-Dame Ouest, H3C 1K3, Montreal, QC, Canada\\
\url{alessandro.koerich@etsmtl.ca}}

\fntext[myfootnote]{Corresponding author. Phone +1 514 396-8574.}

%%%%%%%%%%%%%%%%%%%%%%%%%%%%%%%%%%%%%%%%%%%%%%%%%%%%
\begin{abstract}
Facial expressions are one of the most powerful ways for depicting specific patterns in human behavior and describing human emotional state.
Despite the impressive advances of affective computing over the last decade, automatic video-based systems for facial expression recognition still cannot handle properly variations in facial expression among individuals as well as cross-cultural and demographic aspects. Nevertheless, recognizing facial expressions is a difficult task even for humans.
% Given video sequences captured in-the-wild and a complex emotion representation from dimensional models of affect %(aka, the valence/arousal space), deep FER systems may, however, learn very discriminative feature representations. 
% In this setting, few studies have considered convolutional recurrent neural networks (CRNNs) or 3D-CNNs for spatiotemporal recognition and dimensional representation of emotions.
In this paper, we investigate the suitability of state-of-the-art deep learning architectures based on convolutional neural networks (CNNs) for continuous emotion recognition using long video sequences captured in-the-wild. This study focuses on deep learning models that allow encoding spatiotemporal relations in videos considering a complex and multi-dimensional emotion space, where values of valence and arousal must be predicted. We have developed and evaluated convolutional recurrent neural networks combining 2D-CNNs and long short term- memory units, and inflated 3D-CNN models, which are built by inflating the weights of a pre-trained 2D-CNN model during fine-tuning, using application-specific videos.
Experimental results on the challenging SEWA-DB dataset have shown that these architectures can effectively be fine-tuned to encode the spatiotemporal information from successive raw pixel images and  achieve state-of-the-art results on such a dataset. 
\end{abstract}
%%%%%%%%%%%%%%%%%%%%%%%%%%%%%%%%%%%%%%%%%%%%%%%%%%%%
%%%%%%%%%%%%%%%%%%%%%%%%%%%%%%%%%%%%%%%%%%%%%%%%%%%%

\end{frontmatter}
%\linenumbers

%%%%%%%%%%%%%%%%%%%%%%%%%%%%%%%%%%%%%%%%%%%%
\section{Introduction}
\label{sec:intro}
%%%%%%%%%%%%%%%%%%%%%%%%%%%%%%%%%%%%%%%%%%%%
Facial expressions are the results of peculiar positions and movements of facial muscles over time. According to previous studies, face images and videos provide an important source of information for representing the emotional state of an individual \citep{Survey}. Facial expression recognition (FER) has attracted a growing interest in recent years. The detection of spontaneous facial expressions in-the-wild is a very challenging task, where performance depends on several factors such as variations among individuals, identity bias of subjects such as gender, age, culture and ethnicity, and the quality of recordings (illumination, resolution, head pose, capture conditions). 

Early research on FER systems was inspired by the fundamentals of human affect theory \citep{Ekman1,Ekman2}, where discrete models are employed to classify facial images into discrete categories, such as anger, disgust, fear, happiness, sadness, surprise, that can be recognizable across cultures. The limited generalization capacity of these models has paved the way for multi-dimensional spaces
%(aka valence/arousal space in our case)
that can improve the representativeness and accuracy for describing emotions \citep{EmotionNotUni}.
%Nowadays, however, the generalization capacity  these models have been put into  question, whereas describing emotions into multi-dimensional spaces (aka. valence/arousal space in our case) has shown more representativeness and accuracy in human psychology \citep{EmotionNotUni}.

\begin{figure}[t!]
  \begin{center}
    \includegraphics[width=0.55\linewidth]{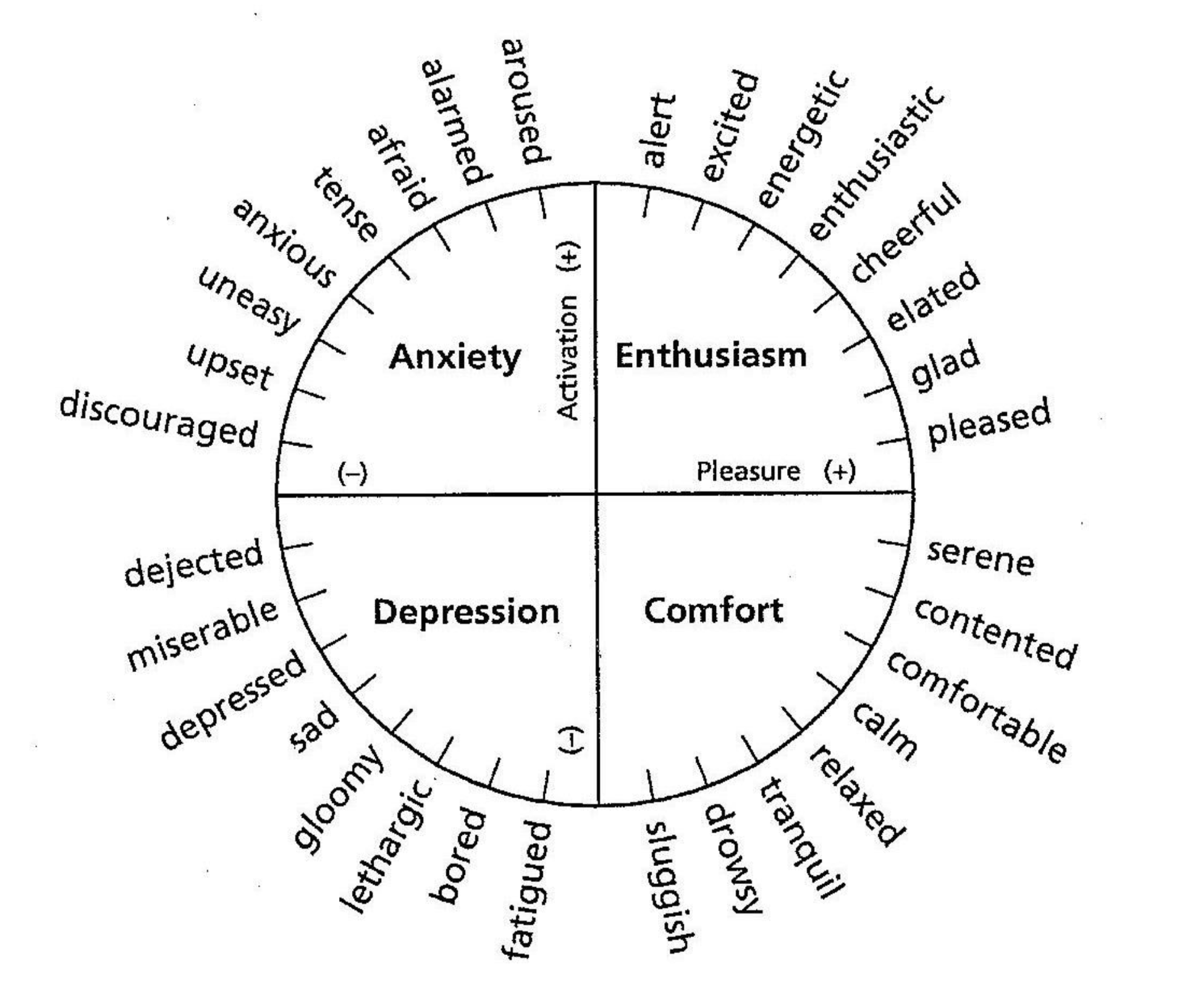}
      \caption{The circumplex model \citep{Circumplex}.}
      \label{fig:circumplex model}
  \end{center}
\end{figure}

There are three levels for describing emotion: pleasantness, attention and levels of activation. Specifically, emotion recognition datasets annotate emotional state with two values named valence and arousal, where the former represents the level of pleasantness and the latter represents the level of activation, each of these values lying into $[-1; 1]$ range. With these values, we are able to project an individual emotional state into a 2D space called the circumplex model, which is shown in Figure~\ref{fig:circumplex model}. The level of arousal is represented on the vertical axis, whereas the level of valence is represented on the horizontal axis.

{Early FER systems relied on shallow approaches, which combined {handcrafted features}, such as {eigenfaces \citep{5661748}, Gabor wavelets \citep{5946775}}, local binary patterns (LBP) \citep{LBP,zavaschi2013fusion} and its variants such as LBP-TOP \cite{4160945} and LBP-SIP \citep{LBP-TOP}, and Weber local descriptor \citep{cossetin2016facial} with support vector machines (SVM). However, recent advances in deep learning (DL) and computing technologies have led FER systems to learn discriminant representations directly from face images. Previous studies, firstly exploited static FER datasets, where subjects are associated to discrete and mutually exclusive categories of emotions. State-of-the-art performance on benchmarking datasets such as FER2013 \citep{FER2013}, TFD \citep{TFD}, and SFEW \citep{SFEW} were achieved using CNN-based approaches \citep{Static1,Static2,Static3,Static4,Static5,Static6,Static7}. }

Temporal information and dynamic facial components can play a crucial role for describing facial expressions \citep{Survey}. By definition, people express emotions in a dynamic process, and learning spatiotemporal structures has become the current trend. Recent studies have proposed deep architectures for FER, which were trained on video data \citep{dynamics}. In the literature, video sequences have been mostly processed using aggregation and concatenation of features for each facial image into clips, yet they cannot leverage the temporal dependencies over a video \citep{Feat.Agg1, Feat.Agg2}. To circumvent this issue, convolutional recurrent neural networks (CRNN) and 3D-CNN architectures have been proposed to encode spatiotemporal relations among video frames \citep{ayral2021temporal, MDN}. 

%%%% CONTRIBUTION %%%%
In this paper, state-of-the-art DL models based on CNNs are investigated and compared for video-based emotion recognition where affective states are represented as continuous values in the bi-dimensional space of valence and arousal. This study focuses on state-of-the-art architectures developed based on pre-trained 2D-CNN that allow encoding spatiotemporal relations of facial features in videos, thereby improving the predictions for continuous values of emotion. Starting from pre-trained 2D-CNNs, we fine-tune two types of DL models with videos: (i) a 2D-CNN combined with a long short -term memory (LSTM) structure; (ii) a 2D-CNN inflated into a 3D-CNN. These models predict emotions through regression of valence and arousal values. We assume that long video sequences are captured in-the-wild, and split each sequence into several clips, not only for augmenting the amount of training data, but also by isolating unique facial expressions.
%(iii) optimizing fine-tuned architectures help model convergence when a low amount of training data is available. 
For proof-of-concept, CNN architectures such as VGG and ResNet have been pre-trained with ImageNet and RAF-DB datasets, and fine-tuned with with multiple video sequences from the SEWA-DB dataset to address the problem of predicting valence and arousal. Experiments were conducted over different clip lengths, overlapping ratios, and strategies for fusing annotations. 
%(iv) the implementation of various initialization and fine-tuning techniques for building and training 3D-CNN models.

This paper is organized as follows. Section~\ref{sec:surv} provides a review of DL models proposed for emotion recognition in videos captured in-the-wild, focusing on models that allow encoding the spatio-temporal relations in facial emotions. Section~\ref{sec:arch} presents an overview of DL models that are proposed for continuous emotion recognition, including pre-processing steps, model architectures, pre-training, fine-tuning procedures and post-processing steps. Section~\ref{sec:res} describes the experimental methodology (e.g., protocol, datasets and performance metrics) used to validate DL models for continuous emotion recognition, as well as the experimental results. Section~\ref{sec:dis} presents a discussion and a comparison with the state-of-the-art. Conclusions are presented in the last section.

%%%%%%%%%%%%%%%%%%%%%%%%%%%%%%%%%%%%%%%%%%%%
\section{Related Work}
\label{sec:surv}
%%%%%%%%%%%%%%%%%%%%%%%%%%%%%%%%%%%%%%%%%%%%

%\subsection{Sequential Face Features Modeling}

A conventional approach for dealing with video frames is to aggregate features extracted from each frame into a clip before final emotion prediction \citet{Feat.Agg2}. In addition to the feature aggregation, \citet{Feat.Agg1} also aggregated mean, variance, minimum and maximum over a sequence of features thus adding some statistical information. However, since feature aggregation cannot exploit inter-correlations between frames and is not able to depict temporal dependencies, this approach has strong limitations. To circumvent this issue, recurrent neural networks (RNN) such as LSTM or 3D-CNN architectures can integrate data series as input, provided that data are sequentially ordered and transitions have a substantial potential of information. While LSTMs can deal with sequential data of variable length in both directions, 3D-CNNs exploit textured variations from sequence of images by extending convolutional kernels to a third dimension. Hence, 3D-CNNs are well suited to encode spatiotemporal information in video-based FER applications. \citet{3D-CNN-2} proposed a 3D-CNN model for action recognition on the UCF101 dataset, which encompasses videos classified over 101 action categories. They have shown that 3D-CNNs can outperform 2D-CNNs on different video analysis benchmarks, and could bring efficient and compact features. Several recent studies \citep{3D-CNN, 3D-CNN-3,3D-CNN-4,3D-CNN-5,3D-CNN-6,3D-CNN-7, Ouyang} have proposed approaches based on 3D-CNNs for FER, nevertheless all of them deal with discrete emotion prediction.

A popular approach for dealing with temporal sequences of frames is a cascaded network, in which architectures {for representation learning and discrimination} are stacked on top of each other, thus various levels of features are leaned by each block and processed by the following until the final prediction. Particularly, the combination of CNNs and LSTM units has been shown effectiveness to learn spatiotemporal representations \citep{3D-CNN-1,3D-CNN-2}. For instance, \citet{Ouyang} used a VGG-16 CNN to extract 16-frame sequence of features and fed an LSTM unit to predict six emotion categories. They pre-processed video frames with a multi-task cascade CNN (MTCNN) to detect faces and described each video by a single 16-frame window. Similarly, \citet{Vielzeuf} used a VGG-16 CNN and an LSTM unit as part of an ensemble with a 3D-CNN, and an audio network. Particularly, they used a method called multi-instance learning (MIL) method to create bag-of-windows for each video with a specific overlapping ratio. Each sequence was described by a single label and contribute to the overall prediction of the matching video clip. 
Since deep neural networks (DNNs) are highly data-dependent, there are strong limitations for designing FER systems based on DNNs, even more since {FER} datasets are often {small and task-oriented} \citep{Survey}. Considering this fact, training deep models on {FER datasets} usually leads to overfitting. In other words, end-to-end training is not feasible if one may learn representation and a discriminant with deep architectures on images with few pre-processing. In this way, some previous work showed that additional task-oriented data for pre-training networks or fine-tuning on well-known pre-trained models could greatly help on building better FER models \citep{PreTrain1,PreTrain5}. Pre-training deep neural networks is then essential for not leading DL models to overfitting. In this way, several state-of-the-art models have been developed and shared for research purposes. VGG-Face~\citep{VGG-FACE} is a CNN based on the VGG-16 architecture \citet{VGG}, with the purpose of circumventing the lack of data by building an architecture for face identification and verification and make it available for the research community. This CNN was trained on about three million images of 2,600 different subjects, which makes this architecture specially adapted both for face and emotion recognition. Recent works that have performed well in FER challenges such as EmotiW \citep{EMOTIW} or AVEC \citep{AVEC} are based on VGG-Face architecture. \citet{VGGFAceSubspace} combined linear discriminant analysis and weighted principal component analysis with VGG-Face for feature extraction and dimension reduction in a face recognition task. \citet{Knyazev}, as part of the EmotiW challenge, fine-tuned a VGG-Face on {the} FER2013 dataset \citep{FER2013}, and aggregated frame features for classifying emotions on video sequences with a linear SVM. Finally, \citet{Ding} proposed peculiar fine-tuning techniques with VGG-Face. They constrained their own network to act like VGG-Face network by transferring the distribution of outputs from late layers rather than transferring the weights.

Recent works used 3D convolutional kernels for spatiotemporal description of visual information. {The first 3D-CNN} models were developed for action recognition task \citep{3D-CNN-1,3D-CNN-2,3D-CNN-5}. 3D-CNNs pre-trained on action recognition datasets were then made available and transferred to affect computing research \citep{3D-CNN-3,3D-CNN-4}. \citet{Ouyang} combined VGG-Face and LSTM among other CNN-RNN and 3D-CNN networks for building an ensemble network for multi-modality fusion (video+audio), which predicts seven categories of emotions. We belied no 3D-CNN {model} has ever been {evaluated} for predicting continuous emotions, and only few approaches {have been} proposed for discrete emotion prediction. This is mainly due to the lack of video datasets for {FER}, which allow exploiting the temporal dimension. To circumvent this issue, \citet{i3D} developed the i3D network, which is able to learn 3D feature representations based on 2D datasets. They inflated a 2D Inception CNN to extend learned weights in 2D to a third dimension. In this way, they developed several pre-trained networks based on the same architecture with combination of ImageNet and Kinetics datasets either on images or optical flow inputs. As demonstrated by \citet{MDN}, convolutional 3D networks (C3D) proposed in previous studies \citep{3D-CNN-3,3D-CNN-4,Ouyang} for emotion recognition and depression detection, has a lower capacity to produce discriminant spatio-temporal features than i3D \citep{i3D}. This is mainly due to the fact that the i3D-CNN is deeper and it benefits of efficient neural connections through the inception module. In this way, with inflated 2D networks we are able to build efficient 3D-CNNs from competitive 2D architectures. \citet{MDN} proposed a deep maximization-differentiation network (MDN) and compared this architecture with i3D-CNN and T-3D-CNN \citep{t3d}, showing that i3D requires fewer parameters than other models and is computationally faster. Finally, \citet{praveen,praveen2} applied i3D networks in the context of pain intensity estimation with ordinal regression. Their approach achieved state-of-the-art results notably by using deep weakly-supervised domain adaptation based on adversarial learning.

{Most of the previous studies on FER} are based on the categorical representation of emotion but some studies {have also dealt with} continuous representations, which have been proved to be effective on both image and video datasets. Discrete models are a very simple representation of emotion and for instance, they do not generalize well across cultures. For instance, smiles can be either attributed to happiness or to fearness or to disgust depending on the context. On the other hand, dimensional models can distinguish emotions upon a better basis, which are levels of arousal and valence \citep{cardinal2015ets}. These two values, widely used in the psychology field, can assign a wider range of emotional states. Researchers have shown that low and high-level features complement each other and their combination could shrink the affective gap, which is defined as the concordance between signal properties or features and the desired output values. For instance, \citet{TWO-STREAM} and \citet{EmotionalMachines}, built feed-forward networks combining color features, texture features {(LBP)} and shape features {(SIFT descriptors)}. Other works focused on emotion recognition {at group level} by studying {not only} facial expressions but also body posture or context \citep{Group}, as well as by exploring various physiological signals such as electrocardiogram and respiration volume \citep{MAHNOB, cardinal2015ets}. \citet{CNN-RNN-VA} compared and used exhaustive variations of CNN-RNN models for valence and arousal prediction on the Aff-Wild dataset \citep{AffWild}. Past studies have particularly worked on full-length short video clips in order to predict a unique categorical label \citep{EMOTIW,AVEC}. However with current datasets and dimensional models, {almost} every frame is annotated and several peaks of emotions can be distinguished \citep{SEWA}. Therefore, a unique label cannot be attributed to a single video clip. The straightforward approach is to split videos into several clips and averaging the predictions on consecutive frames of the sequence to come out at a unique continuous value. Nevertheless, the duration of emotion is not standardized and it is almost totally dependent on random events such as environmental context or subject identity. In this way, windowing video clips is challenging since detecting the most significant sequence for a single unity of emotion is not straightforward. Therefore, fixing arbitrary sequence lengths could bring important biases in emotion prediction and can lead to a loss of information.

%%%%%%%%%%%%%%%%%%%%%%%%%%%%%%%%%%%%%%%%%%%%
\section{Spatiotemporal Models for Continuous Emotion Recognition}
\label{sec:arch}
\begin{figure}[htpb!]
\centering
  \includegraphics[width=0.7\linewidth]{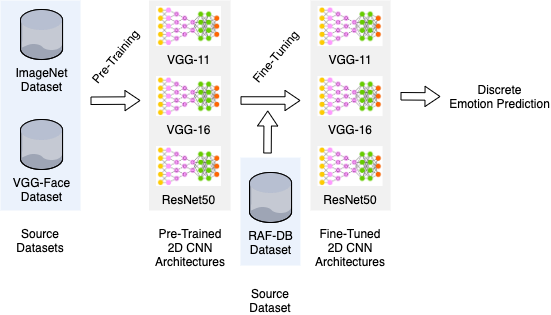}
  \caption{Overview of pre-training and fine-tuning of 2D-CNN architectures for discrete emotion prediction.}
  \label{fig:Over2D}
\end{figure}

%%%%%%%%%%%%%%%%%%%%%%%%%%%%%%%%%%%%%%%%%%%%
{We propose a two-step approach for continuous emotion prediction. In the first step, to circumvent the lack of sequences of continuous labeled videos, we rely on three source image datasets: ImageNet, VGG-Face and RAF-DB. ImageNet and VGG-Face datasets, which contains generic object images and face images, respectively, are used for pre-training three 2D-CNN architectures: VGG-11, VGG-16 and ResNet50. The RAF-DB dataset is closer to the target dataset since it contains face images annotated with discrete (categorical) emotions, and it is used for fine-tuning the 2D-CNN architectures previously trained on ImageNet and VGG-Face datasets, as illustrated in Figure~\ref{fig:Over2D}. Such 2D-CNNs will be used as baseline models with the target dataset.}

\begin{figure}[htpb!]
  \includegraphics[width=\linewidth]{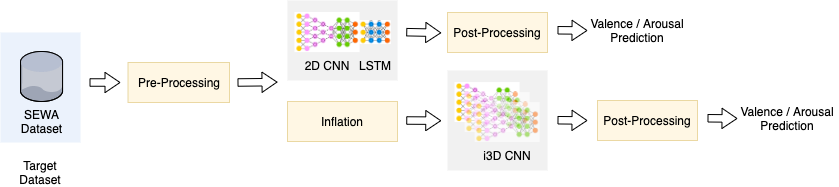}
  \caption{Overview of the two used deep architectures for depicting spatiotemporal features on video sequences: a cascaded network 2D-CNN-LSTM and an i3D-CNN.}
  \label{fig:Over3D}
\end{figure}

{In the second step, we adapt such baseline models for spatiotemporal continuous emotion recognition and we fine-tune them on a target dataset. We use two strategies to model sequential information of videos, as shown in Figure~\ref{fig:Over3D}: (i) a cascade approach where an LSTM unit is added after the last convolutional layer of the 2D-CNNs to form a 2D-CNN-LSTM; (ii) inflating the 2D convolutional layers of the 2D-CNNs to a third dimension to build a i3D-CNN. This second step also includes pre-processing of the videos frames, as well as post-processing of the predictions.}

The rest of this section provides additional information on the pre-training and fine-tuning of 2D-CNN models, the pre-processing steps used to locate face images within video frames and to build the sequences of frames to feed the spatiotemporal models, the DL models for continuous emotion recognition, and post-processing of the emotion predictions.

%%%%%%%%%%%%%%%%%%%%%%%%%%%%%%%%%%%%%%%%%%%%%%%%%%%%%%%%%%%%
\subsection{Pre-Training and Fine-Tuning of 2D-CNNs}
%%%%%%%%%%%%%%%%%%%%%%%%%%%%%%%%%%%%%%%%%%%%%%%%%%%%%%%%%%%%
Training {CNNs} on small datasets systematically leads to overfitting. To {circumvent} this issue, {CNNs can be} pre-trained or fine-tuned on datasets similar {or not} to the target task \citep{PreTrain1,PreTrain5}. Well-known CNN architectures {such as} AlexNet \citep{AlexNet}, VGG~\citep{VGG}, and GoogleNet form an important set of baselines for a large number of tasks, particularly pre-training {such networks} on ImageNet {dataset} constitutes a powerful tool for {representation learning}. {However,} recent FER studies {have shown} that VGG-Face architectures, {which are trained on a very large dataset of face images} overwhelms {architectures trained on} ImageNet for FER applications \citep{VGGFACEbetter}. Furthermore, \citet{Survey} have shown that multi-stage fine-tuning can {provide an} even better performance. We can particularly {mention} FER2013 \citep{FER2013}, TFD \citep{TFD} or more recently RAF-DB \citep{RAF-DB1, RAF-DB2} datasets as good sources of additional data for {FER} tasks. Besides, \citet{tannugi2019memory} and \citet{Cross-dataset} pursued interesting work on cross-dataset generalization task by switching in turn source and target {FER} datasets and evaluating performance of FER models. \citet{Cross-dataset} have shown that datasets are strongly biased and they have developed accordingly, novel architecture that can learn domain-invariant and discriminative features.

Globally, in this study we have considered using three different data sources for double transfer learning \citep{de2019double}: VGG-Face, ImageNet, and RAF-DB. {For the first two datasets, we already have three pre-trained architectures} (VGG-11, VGG-16, ResNet50). {On the other, we had to re-train such architectures on RAF-DB.} We {have evaluated several} configurations for {training and} fine-tuning different CNN architectures with RAF-DB to find out how multi-stage fine-tuning can be well performed. In detail, we fine-tuned CNN architectures by freezing the weights of {certain} early layers while optimizing deeper ones. As architectures are divided into convolution blocks, we have frozen weights according to these blocks. The {proposed} architecture kept convolution blocks {but} classification layers (i.e fully connected layers) were replaced by a stack of {two} fully connected layers with 512 and 128 units respectively, {and an output layer with seven units}, since there are seven different emotion categories in RAF-DB: surprise, fear, disgust, happiness, sadness, anger, and neutral.

%%%%%%%%%%%%%%%%%%%%%%%%%%%%%%%%%%%%%%%%%%%%%%%%%%%%%%%%%%%%%%%
\subsection{Pre-Processing}
%%%%%%%%%%%%%%%%%%%%%%%%%%%%%%%%%%%%%%%%%%%%%%%%%%%%%%%%%%%%%%%
Face images are {usually affected by} background variations such as illumination, head pose, and face patterns linked to some identity bias. In this way, alignment and normalization are the two most commonly used preprocessing methods in face recognition, {which may aid learning discriminant features}. {For instance,} the RAF-DB dataset contains aligned faces, while the subjects in the SEWA-DB dataset are naturally facing a web camera. Then, face alignment is {not} an important issue for this study. Furthermore, normalization only consists in {scaling} pixel values {between} 0 to 1 and to standardize input {dimensions}, faces have been resized to {100$\times$80 pixels, which is the} average dimension of faces {founded in the target dataset}. {On the other hand,} we detail {other} essential steps for face expression recognition in video sequences such as frame and face extraction, {and} window bagging.

%%%%%%%%%%%%%%%%%%%%%%%%%%%%%%%%%%%%%%%%%%%%%%%%%%%%%%%%%%%%%%%
\subsubsection{Frame and Face Extraction}
%%%%%%%%%%%%%%%%%%%%%%%%%%%%%%%%%%%%%%%%%%%%%%%%%%%%%%%%%%%%%%%
The videos of the target dataset (SEWA-DB) have been recorded at 50 frames per second (fps). On the other hand, the valence and arousal annotations are available at each 10 ms, which corresponds to 10 fps. Therefore, it is necessary to replicate annotations for non-labeled frames when using 50 fps.

\begin{figure*}[htpb!]
  \includegraphics[width=\linewidth]{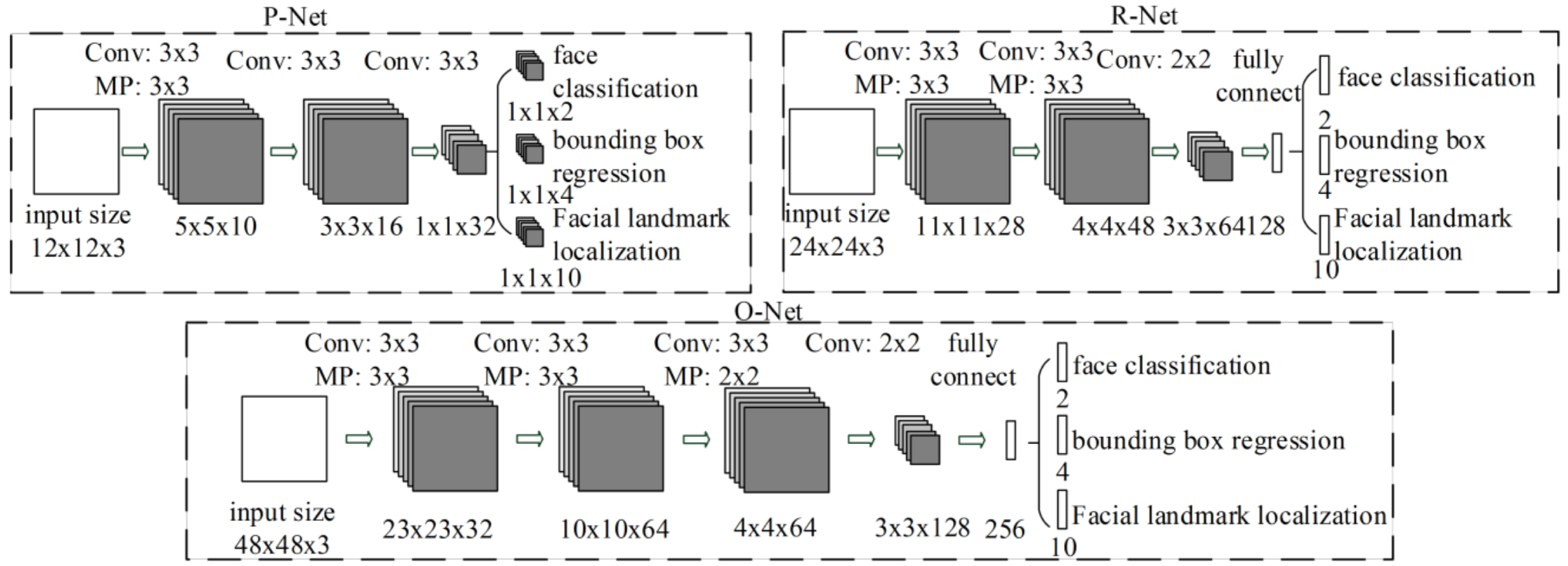}
  \caption{Exhaustive MTCNN architecture composed of a stack of CNNs~\citep{MTCNN}. A single meaningful bounding box and five facial landmarks are extracted from face images.}
  \label{fig:MTCNN}
\end{figure*}

{For locating and extracting faces from the frames of the SEWA-DB videos, we used a multi-task cascaded CNN (MTCNN)~\citep{MTCNN}, which has shown a great efficiency to elect the best bounding box candidates showing a complete face within the image. MTCNN employs three CNNs sequentially to decide which bounding box must be kept according to particular criteria learned by deep learning. The face extractor network outputs box coordinates and five facial landmarks: both eyes, nose and mouth extremities. Once faces are located, they are cropped using the corresponding bounding box. An overview of MTCNN architecture is shown in Figure~\ref{fig:MTCNN}. Only frames showing whole faces are kept, while other frames are discarded.}

%%%%%%%%%%%%%%%%%%%%%%%%%%%%%%%%%%%%%%%%%%%%%%%%%%%%
\subsubsection{Sequence Learning}
%%%%%%%%%%%%%%%%%%%%%%%%%%%%%%%%%%%%%%%%%%%%%%%%%%%%
The target dataset contains long video sequences showing a variety of emotions along records of face expressions from single subjects. Duration of emotions is not clearly established and it varies for each individual. Several studies have been previously {carried out} in order to get expression intensity variations by pointing peak and non-peak expressions along sequences. However, while whole video sequences represent multiple annotations at a specific sampling rate and not a single label, to represent a succession of diverse emotional states, we split the video sequences into several clips of fixed length with a specific overlapping ratio. This has two main advantages: (i) it increases the amount of data for training CNNs; (ii) it allows the investigation of which window settings is better for training spatiotemporal CNNs to learn from long sequences of valence and arousal annotations.
Based on an exploratory study, we have chosen two sequence lengths (16 and 64 consecutive frames of a single video) and three overlapping ratios for each sequence length (0.2, 0.5 and 0.8). For instance, a window of 16 consecutive frames with an overlapping of 0.5 contains the last eight frames of the previous window. 

\begin{figure}[htpb!]
	\centering
		{\includegraphics[width=3.1in]{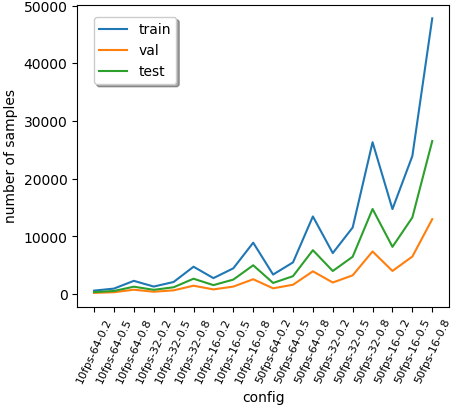}}
		\caption{Evolution of the number of sequences for training, validation and test sets for different configurations: 10 and 50 fps; sequence length of 16, 32 and 64 frames; overlapping ratio of 20\%, 50\% and 80\%.}
    \label{fig:Figure 3} 
\end{figure}

It was also important to check the integrity of contiguous video frames. Indeed some frames are discarded because no face was detected within them, hence damaging the continuity of the temporal information of emotion between each frame. The proposed strategy to divide videos into clips may introduce important temporal gaps between two consecutive frames. Therefore, we applied a tolerance (a temporal difference between close frames) to select clips that give sense to a unique emotion unit. Globally, the MTCNN can detect faces in clips and in average, 90\% of the frames are kept, depending on the sequence length, overlapping ratio and frame rate. Figure~\ref{fig:Figure 3} presents the number of clips available in training, validation and test sets according to such parameters. Finally, the last preprocessing step is to fuse annotations of multiple frames in one clip to get a single emotion label for each window. {For such aim we use either the average of labels or the extremum value of the labels to obtain a single label for each continuous emotion (a single value of valence and a single value of arousal}.

%%%%%%%%%%%%%%%%%%%%%%%%%%%%%%%%%%%%%%%%%%%%%%%%%%%%%%%%%%%%%%%
\subsection{Spatiotemporal Models}
%%%%%%%%%%%%%%%%%%%%%%%%%%%%%%%%%%%%%%%%%%%%%%%%%%%%%%%%%%%%%%%
We {have} developed two spatiotemporal models: (i) a cascaded network based on a VGG-16 network pre-trained on VGG-Face that can be fine-tuned or not on RAF-DB; (ii) an inflated network based on either VGG-11, VGG-16, or ResNet50 architectures pre-trained on different datasets (VGG-Face, RAF-DB, ImageNet).

\subsubsection{Cascaded Networks (2D-CNN-LSTM)}

Long short term memory units (LSTMs) are a special kind of RNN, capable of learning order dependence as we may find in a sequence of frames from a video. The core of LSTMs is a cell state, which adds or removes information depending on the input, output and forget gates. The cell state remembers values over arbitrary time intervals and the gates regulate the flow of input and output information of the cell.

\begin{figure}[htpb!]
	\centering
		\includegraphics[width=5.6in]{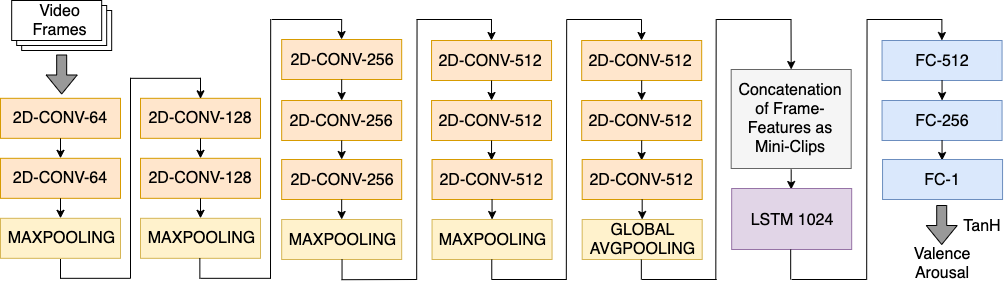}
		\caption{The architecture of the VGG-16-LSTM. Video frames are fed to the CNN and then accumulated at its output to form a feature vector representing one clip. After going through the LSTM unit for modeling temporal information between frames, three fully connected (FC) layers perform the regression of valence and arousal values. For each convolutional layer a 3$\times$3 kernel is used, and the number of filters are indicated. We also detailed the number of units in LSTM and FC layers.}
    \label{fig:Figure 4} 
\end{figure}

The architecture of the proposed cascade network combines the 2D convolutional layers of VGG-16 for representation learning with an LSTM to support sequence prediction, as shown in Figure~\ref{fig:Figure 4}. {The LSTM has a single layer with 1,024 units, with random and uniform distribution initialization to extract temporal features from the face features learned by the 2D-CNN. In order to avoid overfitting, we added some dropout (20\%) and recurrent dropout (20\%) on LSTM units.} Besides, there are also three fully connected layers stacked after the LSTM to improve the expressiveness and accuracy of the model. 

The VGG-16-LSTM architecture is pre-trained considering two different strategies: (i) VGG-16 pre-trained on the VGG-Face dataset; (ii) VGG-16 pre-trained on the VGG-Face dataset and fine-tuned on the RAF-DB dataset. The former strategy adds extra information to the models, such as classification of discrete emotions with RAF-DB, which could help to improve the performance on the regression task. 

%%%%%%%%%%%%%%%%%%%%%%%%%%%%%%%%%%%%%%%%%%%%%%%%%%%%%%%%%%%%%%%
\subsubsection{Inflated 3D-CNN (i3D-CNN)}
\begin{figure}[htpb!]
	\centering
		\includegraphics[width=4in]{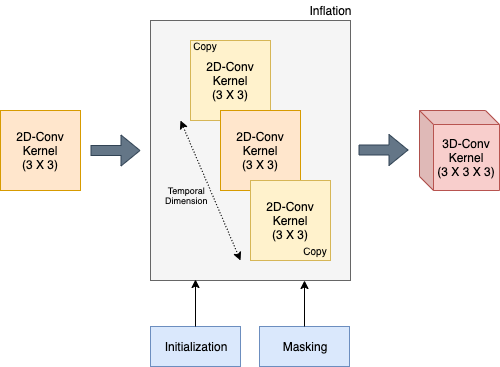}
		\caption{Representation of the inflation method for a single convolutional filter. 2D convolutional kernels are replicated along a new dimension (temporal dimension), to obtain 3D convolutional kernels. Basically, $n\times n$ kernels are made cubic to obtain $n\times n\times n$ kernels. This process is applied to every convolutional filter to transform 2D convolutional layers into 3D convolutional layers.}
    \label{fig:Figure 6} 
\end{figure}

%%%%%%%%%%%%%%%%%%%%%%%%%%%%%%%%%%%%%%%%%%%%%%%%%%%%%%%%%%%%%%%
The need to analyze a sequence of frames led us to the use of 3D-CNNs. 3D-CNNs produce activation maps that allow analyzing data where temporal information is relevant. The main advantage of 3D-CNNs is to learn representation from clips that can strengthen the spatiotemporal relationship between frames. Different from 2D-CNNs, 3D-CNNs are directly trained on batches of frame sequences rather than batches of frames. {On the other hand, adding a third dimension to the CNN architecture increases the number of parameters of the model and that requires much larger training datasets than those required by 2D models. The main downside of using such an architecture for FER tasks is the lack of pre-trained models. Besides that, we cannot consider training 3D-CNN architectures in an end-to-end fashion for continuous emotion recognition due to the limited amount of training data. Therefore, a feasible solution is to resort to weight inflation of 2D-CNN pre-trained models \citep{i3D}. Inflating a 2D-CNN minimizes the need for large amounts of data for training properly a 3D-CNN as the inflation process reuses the weights of the 2D-CNNs.} Figure~\ref{fig:Figure 6} shows that the weight inflation consists of enlarging kernels of each convolution filter by one dimension. Regarding our target task, it means to extend the receptive field of each neuron to the time dimension (a.k.a. a sequence of frames). 2D convolutional kernels are then replicated as many times as necessary to fit the third dimension and form a 3D convolutional kernel. At first glance, pre-trained weights are just copied through the time dimension and provide better approximation for initialization than randomness but do not constitute yet an adequate distribution for the time dimension. With this in mind, the next issue is to find a method that fits best the transfer learning to time dimension with weight inflation by varying some parameters, such as: initialization, masking, multiplier and dilation.

\paragraph{Initialization:} When replicating kernels for weight inflation, it is possible to simply copy the weights $n$ times ($n$ being the dimension of time axis) or to center the weights. Centering means copying once weights of a 2D kernel and initializing the weights of the surrounding kernels that form the 3D filter either randomly (with a uniform distribution) or with zeros. We assume that pre-trained 2D kernels have a good capacity of generalization for images, then giving sufficiently distant distribution for all but one 2D kernel from the copied 2D kernel could have a positive impact on model convergence.
\paragraph{Masking:} Assuming that copied 2D kernels have been pre-trained properly considering  a very similar task and that they perform well on images, the idea of masking is to train adequately inflated weights on time dimension. Then we consider not modifying centered weights during training in order to disseminate the spatial representation learned from pre-trained weights to inflated weights.
\paragraph{Multiplier:} The distribution of CNN weights and the range of targeted values for regression are closely related. Since values of valence and arousal range from $-1$ to $1$ and standard values of the weights often take values between $10^{-3}$ and $10^{-1}$, then rising targeted values by a factor could allow to scale up the distribution space and improve convergence.
\paragraph{Dilation:} As suggested by \citet{Dilation}, we used dilated convolutions on our models. The dilation was performed only on time dimension. We divided the architectures into four blocks with increasing levels of dilation starting from level $1$ for convolutional layers (no dilation) then $2$, $4$ and $8$ for top convolutional layers. Dilated convolution consists of receptive fields larger than conventional ones. In other words, neuron connections of one convolutional layer are spread among neurons of previous layers. Notably, this kind of implementation has shown a good performance for segmentation and object recognition task.

\begin{figure}[htpb!]
	\centering
		\includegraphics[width=5.6in]{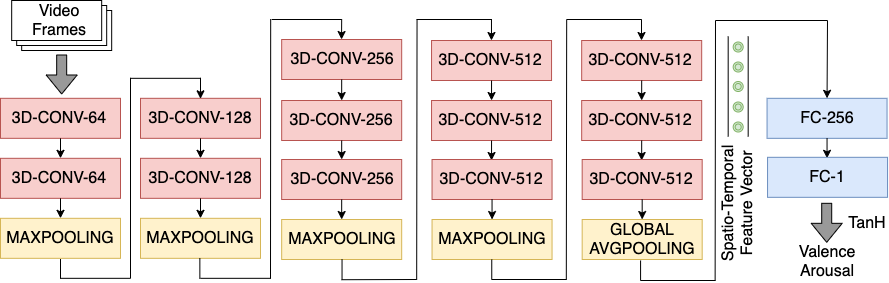}
		\caption{The proposed i3D-CNN based on inflation of 2D convolutional kernels of a pre-trained VGG-16 CNN. Video clips are fed to the i3D-CNN, then spatiotemporal face features are extracted and two fully connected (FC) layers perform the regression of valence and arousal values. For each convolutional layer a 3$\times$3$\times$3 kernel is used, and the number of filters are indicated. We also show the number of units in FC layers.}
    \label{fig:Figure 5} 
\end{figure}

Figure~\ref{fig:Figure 5} shows the architecture of the proposed i3D-CNN, which is based on the inflation of 2D convolutional kernels of a pre-trained VGG-16 CNN. Such a i3D-CNN is then fine-tuned on a target dataset to perform regression of valence and arousal values with a sequence of fully-connected layers.

%%%%%%%%%%%%%%%%%%%%%%%%%%%%%%%%%%%%%%%%%%%%%%%%%%%%%%%%%%%%%%%
\subsection{Post-Processing} 
%%%%%%%%%%%%%%%%%%%%%%%%%%%%%%%%%%%%%%%%%%%%%%%%%%%%%%%%%%%%%%%
The post-processing aims to improve the quality of the prediction by using some statistical information of the target dataset to reduce variance among datasets \citep{ortega2019emotion}. Due to data imbalance in the training set, some values of valence and arousal are difficult to reach. For instance, in the target dataset, neutral emotions, which imply valence and arousal levels close to zero, are much more frequent in the training set than extreme valence and arousal values. We use three post-processing steps: scale normalization, mean filtering, and time delay.

Scale normalization consists in normalizing the predictions according to the distribution of the labels in the training set. Valence and arousal predictions $(y^\prime)$ are normalized by the mean $(\overline{y}_{l_{tr}})$ and the standard deviation $(\sigma_{l_{tr}})$ of the labels of the training set as:

\begin{equation}
\label{norm} 
  y_{sn}=\frac{y^\prime-\overline{y}_{l_{tr}}}{\sigma_{l_{tr}}}
\end{equation}

Mean filtering consists in centering predictions around mean values, increasing the linear relationship and correspondence to the labels. Valence and arousal predictions $(y^\prime)$ are centered by subtracting the mean value of the labels ($\overline{y}_{l_{tr}}$) of the training set and by adding the mean value of the predictions $(\overline{y}^\prime_{tr})$ on the training set as: 

\begin{equation}
\label{mean}
    y_{mf} = y^\prime - \overline{y}_{l_{tr}} + \overline{y}^\prime_{tr}
\end{equation}
    
Finally, time delay is used to compensate some offset between the labels and the predictions due to the reaction-lag of annotators. Valence and arousal predictions $(y^\prime(f))$ at frame $f$ are shifted over $t$ frames (precedent or subsequent) in order to align predictions and labels temporally as:

\begin{equation}
\label{délai}
   {y}_{td}= y^\prime(f+t)
\end{equation}

\noindent where $t$ is an integer in $[-10, 10]$.

%%%%%%%%%%%%%%%%%%%%%%%%%%%%%%%%%%%%%%%%%%%%%%%%%%%%%%%%%%%%%%%
\section{Experimental Results}
\label{sec:res}
%%%%%%%%%%%%%%%%%%%%%%%%%%%%%%%%%%%%%%%%%%%%%%%%%%%%%%%%%%%%%%%
{In this section we present a brief} description of the two FER datasets used in the experiments: RAF-DB and SEWA-DB. Next, we present the performance measures and summarize our experimental setting and the results achieved by the proposed 2D-CNN-LSTM and i3D-CNN models.

%%%%%%%%%%%%%%%%%%%%%%%%%%%%%%%%%%%%%%%%%%%%%%%%%%%%%%%%%%%%%%%
\subsection{Facial Expression Datasets}
%%%%%%%%%%%%%%%%%%%%%%%%%%%%%%%%%%%%%%%%%%%%%%%%%%%%%%%%%%%%%%%
Real World Affective Faces Database (RAF-DB) is a real world dataset that contains 29,672 images downloaded from the Internet \citep{RAF}. Each image has been labeled by around 40 annotators. The dataset has two types of annotation: seven classes of basic emotions and 12 classes of compound emotions. We have only used the seven basic emotions (face images and labels). {Other metadata such as} facial landmarks, bounding box and identity bias such as age, gender, race are also provided but they have not been used {in any step of the proposed approach. RAF-DB was used to fine-tune the pre-trained 2D-CNNs.}

SEWA-DB is a large and richly annotated dataset consisting of six groups of subjects (around 30 people per group), from six different cultural backgrounds (British, German, Hungarian, Greek, Serbian, and Chinese) and divided into pairs of subjects \citep{SEWA}. Each pair had to discuss their emotional state and sentiment toward four adverts previously watched. The dataset consists of 64 videos (around 1,525 minutes of audio visual data), which are split into three folders (34 training, 14 validation, 16 test). Since the labels are not provided for the test set due to its use in FER challenges, we used the validation set as the test set and split the training set into a new training set (28 videos) and validation set (6 videos). Annotations are given for valence, arousal and levels of liking. We only used valence and arousal annotations since previous studies have indicated that the level of liking is not well related with facial expressions.

%%%%%%%%%%%%%%%%%%%%%%%%%%%%%%%%%%%%%%%%%%%%%%%%%%%%%%%%%%%%%%%
\subsection{Performance Metrics}
%%%%%%%%%%%%%%%%%%%%%%%%%%%%%%%%%%%%%%%%%%%%%%%%%%%%%%%%%%%%%%%
{The standard performance metrics used in continuous emotion recognition are the mean absolute error (MAE), the mean absolute percentage error (MAPE), Pearson correlation coefficient (PCC) and concordance correlation coefficient (CCC). PCC assesses distance between target values and predictions and CCC establishes the strength of a linear relationship between two variables. The range of possible values lies in the interval $[-1; 1]$, where $-1$ or $1$ means strong relation and $0$ means no relation at all.} The MAE for a set of labels $y$ and predictions $y^\prime$ is given by:

\begin{equation}
  \text{MAE} = \frac{1}{n}\sum_{i=1}^{n}|y_i^\prime-y_i|  
\end{equation}

The MAPE for a set of labels $y$ and predictions $y^\prime$ is given by:

\begin{equation}
  \text{MAPE} = \frac{1}{n}\sum_{i=1}^{n}\frac{|y_i^\prime-y_i|}{y_i}  
\end{equation}

The PCC is given as:
\begin{equation}
  \rho=\frac{ \sum _{i=1}^{n} \left( y_{i}-\overline{y} \right)  \left( {y^\prime}_{i}-\overline{y}^\prime \right) }{\sqrt{ \sum _{i=1}^{n} \left( y_{i}-\overline{y} \right) ^{2}}~\sqrt{ \sum _{i=1}^{n} \left( {y^\prime}_{i}-\overline{y}^\prime \right) ^{2}}}  
\end{equation}

\noindent where $n$ is the number of samples, $y_{i}$ is the $i$-th label, ${y^\prime}_{i}$ is the $i$-th prediction, and $\overline{y}$ and $\overline{y}^\prime$ are the mean of labels and mean of predictions, respectively.

The CCC combines the PCC with the squared difference between the mean of predictions $\overline{y}^\prime$ and the mean of the labels $\overline{y}$. CCC shows the degree of correspondence between the label and prediction distributions based on the covariance and correspondence. The CCC between a set of labels $y$ and predictions $y^\prime$ is given by:

\begin{equation}
 \rho_c=\frac{2\rho s_ys_{y^\prime}}{s_{y}^{2}+ s_{y^\prime}^{2}+  \left( \overline{y}-\overline{y}^\prime \right) ^{2}}~ 
\end{equation}

\noindent where $s_{y}^{2}$ and $s_{y^\prime}^{2}$ are the variance of $y$ and $y^\prime$ respectively.

%%%%%%%%%%%%%%%%%%%%%%%%%%%%%%%%%%%%%%%%%%%%%%%%%%%%%%%%%%%%%%%
\subsection{Training and Fine-Tuning 2D-CNNs}
%%%%%%%%%%%%%%%%%%%%%%%%%%%%%%%%%%%%%%%%%%%%%%%%%%%%%%%%%%%%%%%
Our first task is to specialize the three pre-trained CNN architectures (VGG-11, VGG-16 and ResNet50) for emotion recognition by fine-tuning them with the RAF-DB dataset. These three architectures were pre-trained either on VGG-Face or ImageNet. For fine-tuning the pre-trained 2D-CNNs on RAF-DB, video frames have been resized to 100$\times$80$\times$3, which is the mean dimension of video frames of the target dataset (SEWA-DB). Learning rate has been fixed to $1e^{-5}$ and batches of size 16. Optimization has been performed with Adam optimizer. We have assigned different weights to each class, according to the number of samples, to deal with the data imbalance found in RAF-DB. This allows that classes with few samples can affect the weights of the model to the same extent as classes with many more samples. Moreover, we have observed that low-level data augmentation such as like rotation, flipping, highlight variations, could help improve the performance. Although data augmentation cannot bring significant information for emotion recognition, it can prevent overfitting on a single sample and improve model distributions.

{The performance achieved by the 2D-CNNs after fine-tuning on RAF-DB is presented in Table~\ref{RAF-DB fine-tuning}, where the suffix BN refers to batch normalization layers added to the original architectures after each convolutional layer to improve model convergence and reduce overfitting. Furthermore, we indicate for each architecture the dataset used for pre-training as well as the convolution block (2\_1 to 5\_1) from which we start fine-tuning the architectures. In general, most of the fine-tuned models achieved accuracy higher than the baseline models \citep{SoA-RAF}. \citet{SoA-RAF} developed two 2D-CNNs to analyze action units detection efficiency on three datasets including RAF-DB. The first CNN was based on residual connections with densely connected blocks (ResNet) and the second architecture was a 2D-CNN consisting of four convolution layers and three fully connected layers. Such baselines achieved 76.54\% and 78.23\% of accuracy on the test set of the RAF-DB dataset, respectively. On the other hand, the proposed VGG-16 CNN model pre-trained with VGG-Face achieved 79.90\% of accuracy on the same test set.}

\begin{table}[htpb!]
\centering
\caption{Results of the fine-tuning of VGG and ResNet50 models on RAF-DB. Convolution block denotes the initial level of fine-tuning of each model. Full denotes that all layers have been fine-tuned.}
\label{RAF-DB fine-tuning}
\footnotesize
\vspace{2mm}
\begin{tabular}{llllc}
\toprule
 & \textbf{} & \textbf{Dataset for} & \multicolumn{1}{l}{\textbf{Convolution}} & \textbf{Accuracy} \\
\textbf{Reference} & \textbf{Model} & \textbf{Pre-Training} & \multicolumn{1}{l}{\textbf{Block}} & \textbf{(\%)} \\
\midrule
\multirow{15}{*}{Proposed} & VGG-11 & ImageNet & \begin{tabular}[l]{@{}l@{}}Full \\ Conv\_2\_1\\ Conv\_3\_1\\ Conv\_4\_1\\ Conv\_5\_1 \end{tabular} & \begin{tabular}[l]{@{}l@{}}75.6\\ 75.9\\ 75.9\\ 75.9\\ 70.3\end{tabular} \\ 
\cmidrule{2-5}
& VGG-11-BN & ImageNet & Full & 77.8 \\ \cmidrule{2-5}
& VGG-16 & VGG-Face & \begin{tabular}[l]{@{}l@{}}Full\\ Conv\_2\_1\\ Conv\_3\_1\\ Conv\_4\_1\\ Conv\_5\_1\end{tabular} & \begin{tabular}[l]{@{}l@{}}78.5\\ 78.5\\ 79.1\\ \bf 79.9\\74.4\\\end{tabular} \\ \cmidrule{2-5}
& VGG-16-BN & VGG-Face & Full & 78.4 \\ \cmidrule{2-5}
& ResNet50 & VGG-Face & \begin{tabular}[l]{@{}l@{}}Full\\ Conv\_4\_1\\ Conv\_5\_1\end{tabular}& \begin{tabular}[l]{@{}l@{}}79.7\\ 78.0\\ 65.1\\\end{tabular} \\ \midrule
\multirow{2}{*}{\citet{SoA-RAF}} & RCNN & NA & NA & 76.5 \\
 & CNN & NA & NA & 78.2 \\ \cmidrule{2-5}
\citet{RAF-SoA3} & ACNN & ImageNet & NA & 85.1 \\
\citet{RAF-SoA} & CNN+RAN & MS-Celeb-1M & NA & 86.9 \\
\bottomrule
\multicolumn{3}{l}{\scriptsize NA: Not applicable.}
\end{tabular}
\end{table}

{Other recent works, which employed} attention networks have achieved better performances \citep{RAF-SoA,RAF-SoA3}. In the proposed approach, we did not have considered two common problems we may found in face analysis in real-world scenarios: occlusions and pose variations. On the contrary, \citet{RAF-SoA} and \citet{RAF-SoA3} addressed these problems by using region-based attention networks. Attention modules are used to extract compact face representations based on several regions cropped from the face and they adaptively adjusts the importance of facial parts. Therefore, these models learn to discriminate occluded and non-occluded faces while improving emotion detection in both cases.

%%%%%%%%%%%%%%%%%%%%%%%%%%%%%%%%%%%%%%%%%%%%%%%%%%%%%%%%%%%%%%%
\subsection{2D-CNN-LSTM Architecture}
%%%%%%%%%%%%%%%%%%%%%%%%%%%%%%%%%%%%%%%%%%%%%%%%%%%%%%%%%%%%%%%
After specializing the three pre-trained CNN architectures (VGG-11, VGG-16 and ResNet50) for emotion recognition by fine-tuning them on the RAF-DB dataset, we develop cascaded networks based on such architectures for spatiotemporal continuous emotion recognition. For such an aim, we have developed two 2D-CNN-LSTM models, one based on the VGG-16 architecture pre-trained on VGG-Face and fine-tuned on RAF-DB because such an architecture achieved the best results on RAF-DB test set, and a second one without fine-tuning on the RAF-DB dataset. Spatial features for each input frame are sequentially provided by the 2D-CNN and the LSTM unit models the temporal information from a single clip. Different configurations were evaluated by varying the length of sequences, the overlapping ratio and the strategy to fuse the labels within a clip. The architectures were fine-tuned on the development set of SEWA-DB and the mean squared error (MSE) was used as cost function. Some other works have also considered CCC as cost function \citep{ortega2019emotion} since it provides information about correspondence and correlation between predictions and annotations. However, we observed a better convergence while using the MSE.

Table~\ref{Tableau 2} shows the results in terms of PCC and CCC considering different frames rates (fps), sequence lengths (SL), overlapping ratios (OR) and fusion modes (FM). In general, both extremum and mean fusion performed well and the best results for both valence and arousal were achieved for sequences of 64 frames at 10 fps. The VGG-16 architecture benefited from fine-tuning on the RAF-DB and it achieved CCC values of 0.625 for valence and 0.557 for arousal on the validation set of SEWA-DB. In addition to the correlation metrics, the proposed 2D-CNN-LSTM achieved an overall MAE of 0.06 (amongst 2D-CNN-LSTM models), which also indicates a good correspondence between predictions and annotations. Since the best performance has been obtained with post-processing steps, thus remodeling our set of predictions and annotations, we have also computed MAPE to evaluate the error ratio between predictions and annotations.  

\begin{table}[htpb!]
\centering
\caption{Performance of the 2D-CNN-LSTM based on the VGG-16 architecture on the SEWA-DB dataset.}
\label{Tableau 2}
\footnotesize
\vspace{2mm}
\begin{tabular}{lcllccc}
\toprule
  \textbf{Dataset for} & \textbf{} & \textbf{} & \textbf{Configuration} & & \\ \cmidrule{2-7}
 \textbf{Initialization} & \textbf{fps} & \textbf{Label} & \textbf{(SL, OR, FM)} & \textbf{PCC$\uparrow$} & \textbf{CCC$\uparrow$} & \textbf{MAPE(\%)$\downarrow$}\\ \midrule 
\multirow{4}{*}{VGG-Face} & \multirow{2}{*}{10} & Valence & 16, 0.2, extremum & 0.590 & 0.560 & 3.8\\
 & & Arousal & 16, 0.2, mean & 0.549 & 0.542 & 8.7\\
\cmidrule{3-7}
  & \multirow{2}{*}{50} & Valence & 64, 0.2, mean & 0.541 & 0.511 & 6.8 \\
 & & Arousal & 64, 0.2, extremum & 0.495 & 0.492 & 3.4\\
 \midrule
   \multirow{4}{*}{RAF-DB} & \multirow{2}{*}{10} & Valence & 64, 0.8, mean & \bf 0.631 & \bf 0.625 & \bf 3.7\\
  & & Arousal & 64, 0.8, extremum & \bf 0.558 & \bf 0.557 & \bf 9.4\\
\cmidrule{3-7}
    & \multirow{2}{*}{50} & Valence & 64, 0.2, mean & 0.582 & 0.568 & 8.2\\
  & & Arousal & 64, 0.2, extremum & 0.517 & 0.517 & 4.4\\
 \bottomrule
\multicolumn{6}{l}{\scriptsize SL: Sequence length (in frames),
OR: Overlapping ratio,
FM: Fusion mode. }\\
\end{tabular}
\end{table}

%%%%%%%%%%%%%%%%%%%%%%%%%%%%%%%%%%%%%%%%%%%%%%%%%%%%%%%%%%%%%%%
\subsection{i3D-CNN Architecture}
%%%%%%%%%%%%%%%%%%%%%%%%%%%%%%%%%%%%%%%%%%%%%%%%%%%%%%%%%%%%%%%
Another alternative for {spatiotemporal modeling} is to use the i3D-CNN. In this way, strong spatiotemporal correlations between frames are directly learned from video clips by a single network. Thanks to weight inflation, we are able to use the pre-trained 2D-CNNs to build i3D-CNN architectures. The inflation method allows us to transpose learned information from various static tasks to dynamic ones, and therefore to perform the essential transfer learning for learning spatiotemporal features. With this in mind, we reused the 2D-CNN architectures shown in Table~\ref{RAF-DB fine-tuning} and expand their convolutional layers to build i3D-CNNs considering two configurations, denoted as C1 and C2 in Table~\ref{Tableau 3}.

Due to the high number of trainable parameters, i3D-CNNs are particularly time-consuming to train and therefore we had to fix the value of some basic hyperparameters instead of performing exploratory experiments to set them. {Therefore, we evaluated only the best configuration found for the 2D-CNN-LSTM, as shown in Table~\ref{RAF-DB fine-tuning}, which uses batch of size 8, sequence length of 64 frames, overlapping ratio of 0.8, and frame rate of 10 fps. This is the main downside of our approach based on i3D CNNs, as the number of trainable parameters of i3D-CNNs is three times greater than the counterpart 2D-CNNs.}

\begin{table}[htpb!]
\centering
\caption{Hyperparameters of i3D-CNN architectures and their possible values.}
\label{Tableau 3}
\footnotesize
\vspace{1mm}
\begin{tabular}{lll}
\toprule
\textbf{Parameters} & \textbf{C1} & \textbf{C2} \\ \midrule
Inflation & Centered & Copied \\ %\toprule
Weight Initialization & Random & Zero \\ %\hline
Masking & No & Yes \\ %\hline
Dilation & \begin{tabular}[c]{@{}c@{}}Bloc1: 1\\ Bloc2: 1\\ Bloc3: 1\\ Bloc4: 1\end{tabular} & \begin{tabular}[c]{@{}c@{}}Bloc1: 1\\ Bloc2: 2\\ Bloc3: 4\\ Bloc4: 8\end{tabular} \\ %\hline
Multiplier & $\times$1 & $\times$100 \\ \bottomrule
\end{tabular}
\end{table}

Tables~\ref{Tableau4} and~\ref{Tableau5} show the best configurations of each architecture for valence and arousal prediction, respectively. Globally, different values of inflation, masking, and dilation did not shown any impact on the results achieved by i3D models. Table~\ref{Tableau6} shows the best performance obtained for each architecture for valence and arousal in terms of PCC and CCC values. Inflated 3D-CNNs for regression seem to be very sensitivity to some configurations for training regarding the range of results achieved by different base models and datasets used in their initialization. In these conditions, it is difficult to state on the effect of a single parameter for inflation. VGG-16 with batch normalization and ResNet50 achieved the best results for both valence and arousal and {have shown a} good ability to predict these values compared to other base models. Surprisingly, the VGG-16 pre-trained on ImageNet achieved higher PCC and CCC for both valence and arousal than those base models pre-trained on VGG-Face and RAF-DB, which are source datasets closer to the target one. On the hand, ResNet50 benefited from the initialization with VGG-Face. In summary, the best results range from 0.313 to 0.406 for PCC and from 0.253 to 0.326 for CCC. These performances still show a poor correlation between predictions and annotations but are comparable to the performance achieved by other studies on continuous emotion prediction that use the SEWA-DB dataset. 

\begin{table}[htpb!]
\centering
\caption{i3D-CNN model configurations according to the best performances for predicting valence.}
\label{Tableau4}
\footnotesize
\vspace{1mm}
\begin{tabular}{lllcclr}
\toprule
 & {\textbf{}} & \multicolumn{5}{c}{\textbf{Parameters}} \\ \cmidrule{3-7} 
 {\textbf{Base}} & {\textbf{Dataset for}} & \textbf{Models} & \textbf{} & \multicolumn{1}{l}{\textbf{}} & \multicolumn{1}{l}{\textbf{Initialization}} & \multicolumn{1}{c}{\textbf{}}\\
 \textbf{Models}& {\textbf{Initialization}} & \textbf{Inflation} & \textbf{Dilation} & \multicolumn{1}{l}{\textbf{Masking}} & \multicolumn{1}{l}{\textbf{Centered Weights}} & \multicolumn{1}{l}{\textbf{Mult}} \\ \midrule
\multirow{2}{*}{VGG-11-BN} & RAF-DB & Centered & I & No & Zero & $\times$1 \\ %\cmidrule{2-7} 
 & ImageNet & Copied & I & No & Zero & $\times$100 \\ \cmidrule{2-7}
\multirow{3}{*}{VGG-16} & VGG-Face & Centered & I & No & Random & $\times$1 \\ %\cmidrule{2-7} 
 & RAF-DB & Copied & I & No & Random & $\times$1 \\ %\cmidrule{2-7} 
 & ImageNet & Centered & I & No & Random & $\times$1 \\ \cmidrule{2-7}
\multirow{3}{*}{VGG-16-BN} & VGG-Face & Centered & I & No & Random & $\times$1 \\ %\cmidrule{2-7} 
 & RAF-DB & Copied & I & Yes & Random & $\times$1 \\ %\cmidrule{2-7} 
 & ImageNet & Copied & I & No & Random & $\times$1 \\ \cmidrule{2-7}
\multirow{3}{*}{ResNet50} & VGG-Face & Centered & I & Yes & Zero & $\times$100 \\ %\cmidrule{2-7} 
 & RAF-DB & Copied & VIII & No & Zero & $\times$1 \\ %\cmidrule{2-7} 
 & ImageNet & Centered & VIII & Yes & Zero & $\times$1 \\ \bottomrule
\end{tabular}
\end{table}

\begin{table}[htpb!]
\centering
\caption{i3D-CNN model configurations according to the best performances for predicting arousal.}
\label{Tableau5}
\footnotesize
\vspace{1mm}
\begin{tabular}{lllcclr}
\toprule
 & {\textbf{}} & \multicolumn{5}{c}{\textbf{Parameters}} \\ \cmidrule{3-7} 
 {\textbf{Base}} & {\textbf{Dataset for}} & \textbf{} & \textbf{} & \multicolumn{1}{l}{\textbf{}} & \multicolumn{1}{l}{\textbf{Initialization}} & \multicolumn{1}{c}{\textbf{}}\\
 \textbf{Models} & {\textbf{Initialization}} & \textbf{Inflation} & \textbf{Dilation} & \multicolumn{1}{l}{\textbf{Masking}} & \multicolumn{1}{l}{\textbf{Centered Weights}} & \multicolumn{1}{l}{\textbf{Mult}} \\ \midrule
\multirow{2}{*}{VGG-11-BN} & RAF-DB & Centered & I & No & Random& $\times$1 \\ %\cmidrule{2-7} 
 & ImageNet & Centered & VIII & No & Zero & $\times$1 \\ \cmidrule{2-7}
\multirow{3}{*}{VGG-16} & VGG-Face & Centered & I & No & Random & $\times$1 \\ %\cmidrule{2-7} 
 & RAF-DB & Copied & I & Yes & Random& $\times$100 \\ %\cmidrule{2-7} 
 & ImageNet & Centered & I & No & Random& $\times$1 \\ \cmidrule{2-7}
\multirow{3}{*}{VGG-16-BN} & VGG-Face & Centered & I & Yes & Random& $\times$1 \\ %\cmidrule{2-7} 
 & RAF-DB & Copied & I & No & Random & $\times$1 \\ %\cmidrule{2-7} 
 & ImageNet & Centered & I & No & Zero & $\times$1 \\ \cmidrule{2-7}
\multirow{3}{*}{ResNet50} & VGG-Face & Copied & I & Yes & Zero & $\times$100 \\ %\cmidrule{2-7} 
 & RAF-DB & Centered & VIII & Yes & Zero & $\times$1 \\ %\cmidrule{2-7} 
 & ImageNet & Centered & I & No & Zero & $\times$1 \\ \bottomrule
\end{tabular}
\end{table}

\begin{table}[htpb!]
\centering
\caption{Best performances of i3D-CNNs for predicting valence and arousal in terms of PCC and CCC values and MAPE according to different models and their initialization.}
\label{Tableau6}
\footnotesize
\vspace{1mm}
\begin{tabular}{llcccccc}
\toprule %\hline
 \textbf{Base}& {\textbf{Dataset for}} & \multicolumn{3}{c}{\textbf{Valence}} & \multicolumn{3}{c}{\textbf{Arousal}} \\ \cmidrule{3-5} \cmidrule{6-8} 
 {\textbf{Models}} & {\textbf{Initialization}} & \textbf{PCC$\uparrow$} & \textbf{CCC$\uparrow$} & \textbf{MAPE(\%)$\downarrow$} & \textbf{PCC$\uparrow$} & \textbf{CCC$\uparrow$} & \textbf{MAPE(\%)$\downarrow$}\\  \midrule 
\multirow{2}{*}{VGG-11-BN} & RAF-DB & 0.035 & 0.018 & 2.9 & 0.359 & 0.348 & 8.1\\ %\cmidrule{2-6} 
& ImageNet & 0.040 & 0.025 & 4.2 & 0.342 & 0.203 & 3.0\\ \cmidrule{2-8}
\multirow{3}{*}{VGG-16} & VGG-Face & 0.119 & 0.071 & 3.0 &  0.220 & 0.166 & 5.2 \\
 & RAF-DB & 0.036 & 0.028 & 3.5 & 0.242 & 0.119 & 4.6\\ %\cmidrule{2-6} 
 & ImageNet & 0.209 & 0.190 & 2.4 & 0.391 & 0.189 & 3.8\\ \cmidrule{2-8}
\multirow{3}{*}{{VGG-16-BN}} & VGG-Face & 0.203 & 0.105 & 3.6 & 0.347 & 0.304 & 5.3\\ %\cmidrule{2-6} 
 & RAF-DB & 0.123 & 0.101 & 3.2 & 0.284 & 0.165 & 3.3\\ %\cmidrule{2-6} 
 & {ImageNet} & \textbf{0.346} & \textbf{0.304} & \textbf{5.6} & \textbf{0.382} & \textbf{0.326} & \textbf{5.3}\\ \cmidrule{2-8}
\multirow{3}{*}{{ResNet50}} & {VGG-Face} & \textbf{0.313} & \textbf{0.253} & \textbf{3.5}& \textbf{0.406} & \textbf{0.273} & \textbf{4.9}\\ %\cmidrule{2-6} 
 & RAF-DB & 0.113 & 0.063 & 2.9 & 0.262 & 0.207 & 4.9\\ %\cmidrule{2-6} 
 & ImageNet & 0.183 & 0.164 & 6.0 & 0.323 & 0.256 & 4.7\\ \bottomrule
\end{tabular}
\end{table}

%\newpage
%%%%%%%%%%%%%%%%%%%%%%%%%%%%%%%%%%%%%%%%%%%%%%%%%%%%%%%%%%%%%%%
\section{Discussion}
\label{sec:dis}
%%%%%%%%%%%%%%%%%%%%%%%%%%%%%%%%%%%%%%%%%%%%%%%%%%%%%%%%%%%%%%%
The experiments carried out on SEWA-DB have shown that the 2D-CNN-LSTM architectures achieved (Table~\ref{Tableau 2}) better results than i3D-CNN architectures (Table~\ref{Tableau6}). Notably, for the former, valence was better predicted than arousal in terms of CCC. On the contrary, for the latter, arousal was better predicted than valence, also in terms of CCC. Previous works have raised the fact that intuitively face textures on video sequences are the main source of information for describing the level of positivity in emotions, hence valence values. In contrast, arousal is better predicted with voice frequencies and audio signals. However, our work with inflated networks suggests that simultaneous learning spatiotemporal face features benefits the prediction of arousal values. 
%Moreover, we observed a relatively low MAPE of 5.3\% both for predicting valence and arousal and for our best models, which is quite higher than for the detection of valence (3.7\%) and quite lower than for the detection of arousal (9.4\%) on our best 2D-CNN-LSTM models. 

Regarding the complexity of the two proposed approaches for continuous emotion recognition, we had to make some trade-off that certainly have impacted the quality of the results provided by i3D-CNN architectures. We have also observed a high sensitivity in the training of this type of architecture according to various configurations. This implies that i3D-CNN architectures are very flexible and further improvement could lie in better initialization and tuning of the number and quality of parameters regarding the potential of this model. Furthermore, inflated weights provided good initialization for action recognition tasks which suggested that we could also take advantage of this method for emotion recognition. However, the main difference is that for action recognition, researchers had hundreds of various short videos for a classification task while we have relatively long videos of few subjects for a regression task. Nevertheless, the experimental results have shown a great potential for further improvement if more data is available for fine-tuning the i3D models. On the other hand, the performance of 2D-CNN-LSTM architectures were very satisfying and this type of architecture is still a good choice for FER applications.  

\begin{table}[htpb!]
\centering
\caption{Comparison of the best results achieved by the proposed models and the state-of-the-art for SEWA-DB dataset.}
\label{T4.3}
\footnotesize
\vspace{2mm}
\begin{tabular}{l l c c c c}
\toprule %\hline
&  & \multicolumn{2}{c}{\textbf{Valence}} & \multicolumn{2}{c}{\textbf{Arousal}} \\ \cmidrule{3-4}
\cmidrule{5-6} 
 {\textbf{Reference}} & {\textbf{Model}} & \textbf{PCC$\uparrow$} & \textbf{CCC$\uparrow$} & \textbf{PCC$\uparrow$} & \textbf{CCC$\uparrow$} \\ \midrule 
\multirow{2}{*}{Proposed} & 2D-CNN-LSTM$^1$  & \bf 0.631 & \bf 0.625 & \bf 0.558 & \bf 0.557 \\ 
 & i3D-CNN$^2$  & {0.346} & {0.304} & {0.382} & {0.326} \\
 \cmidrule{2-6}
\multirow{5}{*}{\cite{SEWA}}& SVR & 0.321 & 0.312 & 0.182 & 0.202 \\ 
& RF & 0.268 & 0.207 & 0.181 & 0.123 \\ 
& LSTM & 0.322 & 0.281 & 0.173 & 0.115 \\ 
& ResNet18 (RMSE) & 0.290 & 0.270 & 0.130 & 0.110 \\ 
& {ResNet18 (CCC)} & {0.350} & {0.350} & {0.350} & {0.290} \\
\midrule 
\cite{10.1145/3347320.3357690}$^\star$& ST-GCN & NA & 0.540 & NA & 0.581 \\
\cite{10.1145/3347320.3357692}$^\star$& DenseNet-style CNN & NA & 0.580 & NA & 0.594 \\
\bottomrule
\multicolumn{6}{l}{\scriptsize $^1$VGG-16 fine-tuned with RAF-DB. $^2$Inflated VGG-16-BN, initialized with ImageNet.}\\
\multicolumn{6}{l}{\scriptsize NA: Not Available. $^\star$Results are reported for a subset of SEWA-DB encompassing only}\\
\multicolumn{6}{l}{three cultures (Hungarian, German, Chinese).}
\end{tabular}
\end{table}

Table~\ref{T4.3} shows the best results achieved by the proposed 2D-CNN-LSTM and i3D-CNN models and compare them with the baseline models proposed by \citet{SEWA}. For ResNet18, they have evaluated Root Mean Squared Error (RMSE) and CCC as loss function. The CCCs achieved by the i3D-CNN are slightly higher than those achieved by all models of \citet{SEWA}. On the other hand, the CCC values achieved by the 2D-CNN-LSTM are almost twice than the best results achieved by the best model (ResNet18) of \citet{SEWA}. Table~\ref{T4.3} also shows the results achieved by \cite{10.1145/3347320.3357690} and \cite{10.1145/3347320.3357692}, which are not directly comparable since both used a subset of SEWA-DB encompassing only three cultures. They optimized emotion detection for two cultures (Hungarian, German) to perform well on the third one (Chinese). Notably, \cite{10.1145/3347320.3357690} proposed a combination of 2D-CNN and 1D-CNN, which has fewer parameters than 3D-CNNs, and a spatiotemporal graph convolution network (ST-GCN) to extract appearance features from facial landmarks sequences. \cite{10.1145/3347320.3357692} used a VGG-style CNN and a DenseNet-style CNN to learn cross-culture face features, which were used to predict two adversarial targets: one for emotion prediction, another for culture classification. In conclusion, these two methodologies achieved state-of-the-art results on cross-cultural emotion prediction tasks.

%%%%%%%%%%%%%%%%%%%%%%%%%%%%%%%%%%%%%%%%%%%%%%%%%%%%%%%%%%%%%%%
\section{Conclusion}
\label{sec:conc}
%%%%%%%%%%%%%%%%%%%%%%%%%%%%%%%%%%%%%%%%%%%%%%%%%%%%%%%%%%%%%%%
{In this paper, we have presented two CNN architectures for continuous emotion prediction in-the-wild. The first architecture is a combination of a fine-tuned VGG-16 CNN and an LSTM unit. Such an architecture achieved state-of-the-art results on the SEWA-DB dataset, producing CCC values of 0.625 and 0.557 for valence and arousal prediction, respectively. The second architecture is based on the concept of inflation, which transfers knowledge from pre-trained 2D-CNN models into a 3D to model temporal features. The best proposed i3D-CNN architecture achieved CCC values of 0.304 and 0.326 for valence and arousal prediction, respectively. These values are far below than those achieved by the 2D-CNN-LSTM. Due to the high number of parameters of i3D-CNNs (barely 3 times greater than 2D-CNNs), fine-tuning and hyperparameter tuning of such architectures require a huge computational effort as well as huge datasets. Unfortunately, facial expression datasets for continuous emotion recognition are relatively small for such a task.}

{We have also shown that a double transfer learning strategy over VGG and ResNet architectures with ImageNet and RAF-DB datasets can improve the accuracy of the baseline models. It should be noticed that in this work, subjects were mostly facing the camera with relative clear view of the whole face. To some extent, this could imply some bias in the results when presenting diverse real world scenarios. Moreover, the complexity of the i3D-CNN architecture could at this time be a drag for live applications. Finally, to the best of our knowledge, it was the first time that 3D-CNNs were used in regression applications for predicting valence and arousal values for emotion recognition.}

There are some promising directions to expand the approaches proposed in this paper. One could take advantage of the development of huge and complex cross-cultural datasets such as the Aff-Wild dataset to exploit occlusion cases, pose variations or even scene breaks. In particular, with i3D-CNN architectures, we believe deep learning algorithms possess the capacity and robustness to deal with these specific cases and benefit from an adequate flexibility to analyze both the separability and combination of discriminant spatiotemporal features.
%Thus, having promising results for facial emotion recognition on video sequences.
Finally, we have shown a peculiar and flexible way of fine-tuning inflated CNNs and maybe this strategy could be transferred to other applications such as object and action recognition on video sequences.

%%%%%%%%%%%%%%%%%%%%%%%%%%%%%%%%%%%%%%%%%%%%%%%%%%%%%%%%%%%%%%%
 %\section*{Availability of Data and Material}
%%%%%%%%%%%%%%%%%%%%%%%%%%%%%%%%%%%%%%%%%%%%%%%%%%%%%%%%%%%%%%%

%%%%%%%%%%%%%%%%%%%%%%%%%%%%%%%%%%%%%%%%%%%%%%%%%%%%%%%%%%%%%%%
\section*{Competing Interests}
The authors declare that they have no competing interests.
%%%%%%%%%%%%%%%%%%%%%%%%%%%%%%%%%%%%%%%%%%%%%%%%%%%%%%%%

%%%%%%%%%%%%%%%%%%%%%%%%%%%%%%%%%%%%%%%%%%%%%%%%%%%%%%%%%%%%%%%
%\section*{Acknowledgements}
%We would like to thank O. M. Parkhi, A. Vedaldi, and Andrew Zisserman, Q. Cao et al., for their work on face recognition with VGG-Face and VGG-Face 2 \citep{Model1,VGG-FACE} datasets and the availability of their pre-trained networks which were very helpful for our study. 
%%%%%%%%%%%%%%%%%%%%%%%%%%%%%%%%%%%%%%%%%%%%%%%%%%%%%%%%%%%%%%%

%%%%%%%%%%%%%%%%%%%%%%%%%%%%%%%%%%%%%%%%%%%%%%%%%%%%%%%%%%%%%%%
%\section*{References}
%%%%%%%%%%%%%%%%%%%%%%%%%%%%%%%%%%%%%%%%%%%%%%%%%%%%%%%%%%%%%%%
\bibliography{refs}

\end{document}